\documentclass{IEEEtaes}

\usepackage{color,array,amsthm}
\usepackage{graphicx}
\usepackage{amsmath,amsfonts,amssymb}
\usepackage{algorithmic}
\usepackage{algorithm}
\usepackage{cite}
\usepackage{booktabs}
\usepackage{hyperref}
\usepackage{url}
\usepackage{multirow}

\jvol{XX}
\jnum{XX}
\jmonth{XXXXX}
\paper{1234567}
\pubyear{2026}
\doiinfo{TAES.2026.Doi Number}

\newcommand{\bbbeta}{\boldsymbol{\beta}}
\newcommand{\bbmu}{\boldsymbol{\mu}}

\newcommand{\bbSig}{\boldsymbol{\Sigma}}
 
\newcommand{\bbphi}{\boldsymbol{\phi}} 
\newcommand{\bx}{\mathbf{x}}
\newcommand{\by}{\mathbf{y}}
\newcommand{\bz}{\mathbf{z}}
\newcommand{\bu}{\mathbf{u}}
\newcommand{\bp}{\mathbf{p}}
\newcommand{\calD}{\mathcal{D}}
\newcommand{\calF}{\mathcal{F}}

\newcommand{\calN}{\mathcal{N}}

\newcommand{\calJ}{\mathcal{J}}
\newcommand{\Real}{\mathbb{R}}

\begin{document}

\title{Diffusion Policy with Bayesian Expert Selection for Active Multi-Target Tracking}

\author{HAOTIAN XIANG}
\affil{School of ECE, University of Georgia, Athens, GA 30602, USA}

\author{QIN LU}
\affil{School of ECE, University of Georgia, Athens, GA 30602, USA}

\author{YAAKOV BAR-SHALOM}
\affil{Dept. of ECE, University of Connecticut, Storrs, CT 06269, USA}


\corresp{{\itshape (Corresponding author: Qin Lu)}}

\authoraddress{H. Xiang and Q. Lu are with the School of Electrical and Computer Engineering, University of Georgia, Athens, GA 30602 USA. Y. Bar-Shalom is with the Department of Electrical and Computer Engineering, University of Connecticut, Storrs, CT 06269 USA. H. Xiang and Q. Lu are supported by NSF \# 2340049.}

\markboth{XIANG ET AL.}{DIFFUSION POLICY WITH BAYESIAN EXPERT SELECTION FOR ACTIVE MULTI-TARGET TRACKING}
\maketitle

\begin{abstract}
Active multi-target tracking requires a mobile robot to balance exploration for undetected targets with exploitation of uncertain tracked ones. Diffusion policies have emerged as a powerful approach for capturing diverse behavioral strategies by learning action sequences from expert demonstrations. However, existing methods implicitly select among strategies through the denoising process, without uncertainty quantification over which strategy to execute. We formulate expert selection for diffusion policies as an offline contextual bandit problem and propose a Bayesian framework for pessimistic, uncertainty-aware strategy selection. A multi-head Variational Bayesian Last Layer (VBLL) model predicts the expected tracking performance of each expert strategy given the current belief state, providing both a point estimate and predictive uncertainty. Following the pessimism principle for offline decision-making, a Lower Confidence Bound (LCB) criterion then selects the expert whose worst-case predicted performance is best, avoiding overcommitment to experts with unreliable predictions. The selected expert conditions a diffusion policy to generate corresponding action sequences. Experiments on simulated indoor tracking scenarios demonstrate that our approach outperforms both the base diffusion policy and standard gating methods, including Mixture-of-Experts selection and deterministic regression baselines.
\end{abstract}

\begin{IEEEkeywords}Active target tracking, diffusion policy, Bayesian inference, contextual bandits, uncertainty quantification, mixture of experts
\end{IEEEkeywords}

\section{INTRODUCTION}
\label{sec:introduction}

Active multi-target tracking requires a mobile robot to autonomously navigate and maintain accurate state estimates of multiple moving targets under limited field-of-view (FoV) constraints~\cite{sun2024moving,hero2011sensor, bostrom2021sensor, bostrom2021pmbm}. A fundamental challenge is the exploration-exploitation dilemma: the robot must decide whether to explore unobserved regions to discover new or lost targets, or to exploit current detections by following tracked targets to reduce estimation uncertainty~\cite{liu2025matt,schlotfeldt2018anytime}. This challenge is inherently multi-modal: different situations call for fundamentally different behavioral strategies, and the optimal strategy depends critically on the robot's current belief state. Traditional approaches, from pursuit-evasion formulations~\cite{lavalle1997motion, ragi2013uav} to information-theoretic planning~\cite{le2009trajectory,hero2011sensor} and deep reinforcement learning~\cite{jeong2021deep,yang2023policy}, typically produce unimodal policies that either average across modes or collapse to a single dominant strategy. Standard regression-based imitation learning similarly suffers from mode-averaging when demonstrations contain qualitatively different behaviors~\cite{pearce2023imitating}. Moreover, optimization- and sampling-based planners incur online replanning costs that limit decision frequency~\cite{sudha2026informative}, motivating learned policies that amortize planning costs through efficient inference~\cite{lew2025aid,karaman2011sampling}. Our work addresses this gap by learning to select among pre-trained expert planners with distinct behavioral strategies.

Diffusion policies~\cite{chi2025diffusion,ho2020denoising} have provided a powerful solution by representing policies as conditional denoising processes that naturally capture multi-modal action distributions, with success across manipulation~\cite{chi2025diffusion,ze20243d}, navigation~\cite{sridhar2024nomad,cao2025dare}, and foundation models for robotics~\cite{liu2024rdt,team2024octo}. In active tracking, MATT-Diff~\cite{liu2025matt} learns a diffusion policy from demonstrations generated by multiple expert planners with distinct exploration-exploitation behaviors. However, while diffusion policies excel at representing \textit{what} behavioral modes exist, they provide limited mechanisms for deciding \textit{when} each mode should be executed. In MATT-Diff, mode selection is implicitly determined by random noise in the denoising process, lacking any principled mechanism for uncertainty-aware strategy selection based on the current belief state~\cite{liu2025matt}.

Recent work has begun to address explicit mode selection through Mixture of Experts (MoE) architectures in diffusion models~\cite{reuss2024efficient,zhou2024variational,wang2024sparse}, where a softmax gating network routes inputs to specialized experts. However, softmax gating produces point-estimate routing probabilities without uncertainty quantification, making it difficult to assess whether a given expert assignment is reliable or whether the selector is extrapolating beyond its training distribution. The broader MoE literature has focused on improving routing quality through load balancing~\cite{shazeer2017outrageously,fedus2022switch} and expert specialization~\cite{hendawy2023multi}, but all operate within this classification-based paradigm without associated uncertainty.

We propose an alternative perspective: framing expert selection as an offline contextual bandit problem~\cite{li2010contextual,russo2018tutorial}, where a regression model predicts each expert's expected tracking reward given the current belief state, and the expert with the best predicted outcome is selected. This formulation offers two practical advantages. First, by predicting each expert's expected performance as a continuous reward, the offline setting calls for \textit{pessimism in the face of uncertainty}~\cite{nguyen2021offline}, a principle well-established in offline reinforcement learning and bandits, where the agent should be conservative about experts whose predicted rewards are uncertain. We instantiate this via a Lower Confidence Bound (LCB) criterion that penalizes high predictive variance, naturally avoiding overcommitment to experts with poor training coverage. Second, the regression setting pairs naturally with efficient Bayesian last-layer methods: we adopt Variational Bayesian Last Layers (VBLL)~\cite{harrison2024variational}, which provide predictive uncertainty in a single forward pass without requiring ensembles or multiple stochastic passes. The reward model is trained offline from demonstration data and held fixed at deployment: LCB leverages predictive uncertainty to conservatively select experts, preferring those whose predicted rewards are both high and confident.

Our contributions are as follows:
\begin{itemize}
    \item We identify a key limitation in current diffusion policies for multi-modal tasks: the reliance on stochastic sampling for implicit mode selection, without mechanisms for uncertainty-aware strategy adaptation.
    \item We formulate expert selection for diffusion policies as an offline contextual bandit problem, thereby bridging the offline bandit and diffusion policy literatures and providing a principled framework for uncertainty-aware strategy routing.
    \item We develop a practical algorithm combining VBLL for performance prediction with a LCB criterion for pessimistic expert selection. Experiments on multi-target tracking demonstrate that our approach outperforms both the base diffusion policy and standard MoE gating methods.
\end{itemize}

\section{RELATED WORK}
\label{sec:related_work}

\subsection{Active Target Tracking}
\label{subsec:related_tracking}

Active target tracking spans pursuit-evasion games~\cite{lavalle1997motion}, information-theoretic planning~\cite{le2009trajectory,hero2011sensor}, and multi-robot coordination~\cite{zhou2018resilient,schlotfeldt2018anytime, zhang2025cooperative,duan2020dynamic}. Deep reinforcement learning approaches learn end-to-end tracking policies~\cite{jeong2021deep,yang2023policy}, though these typically learn a single behavioral mode and do not address the problem of selecting among qualitatively different strategies at deployment. Learning from multiple expert planners requires a policy representation that can capture diverse behavioral modes without mode-averaging, motivating the use of diffusion-based generative models.

\subsection{Diffusion Policies for Robot Learning}
\label{subsec:related_diffusion}

Diffusion Policy~\cite{chi2025diffusion} represents robot policies as conditional denoising processes~\cite{ho2020denoising}, naturally capturing multi-modal action distributions and addressing the mode-averaging problem of regression-based behavior cloning~\cite{pearce2023imitating}. For active tracking, MATT-Diff~\cite{liu2025matt} trains a diffusion policy on demonstrations from multiple expert planners, successfully learning multi-modal action distributions. However, the behavioral mode executed at each step is determined by the stochastic denoising process rather than by an explicit selection mechanism. Our work decouples mode selection from action generation by introducing a separate, uncertainty-aware expert selector that conditions the diffusion policy.

\subsection{Mixture of Experts in Diffusion Models}
\label{subsec:related_moe}

MoE architectures route inputs to specialized sub-networks via a gating network. In diffusion models, MoE has been applied for temporal specialization across denoising timesteps~\cite{reuss2024efficient}, policy distillation with fewer denoising steps~\cite{zhou2024variational}, and multi-task robotic manipulation conditioned on task identifiers~\cite{wang2024sparse}. In all cases, the gating function is a standard softmax network producing point-estimate routing probabilities. While prior work addresses routing quality through load-balancing losses~\cite{shazeer2017outrageously,fedus2022switch} and expert specialization~\cite{hendawy2023multi}, all operate within a classification-based paradigm. Our work addresses a complementary question: \textit{how confident is the routing decision}, which requires uncertainty quantification that softmax gating does not provide.

\subsection{Uncertainty Quantification for Neural Networks}
\label{subsec:related_uq}

Uncertainty quantification (UQ) for neural networks spans a spectrum from full Bayesian treatment to lightweight post-hoc approximations~\cite{blundell2015weight,lakshminarayanan2017simple}. Monte Carlo (MC) Dropout reinterprets dropout at test time as approximate variational inference~\cite{gal2016dropout} but often exhibits poor calibration under distribution shift~\cite{ovadia2019can}. Bayesian Last Layer (BLL) methods offer a practical middle ground by restricting Bayesian inference to the final layer while using deterministic features from earlier layers~\cite{watson2021latent}. VBLL extends BLL with variational inference over a structured Gaussian posterior on last-layer weights, enabling end-to-end ELBO training with demonstrated calibration benefits~\cite{harrison2024variational, xiang2025fine, xiang2026scalable}. We adopt VBLL because our expert selector requires single-pass uncertainty estimates for real-time robot control, a constraint that excludes methods requiring multiple forward passes, such as MC Dropout.

\subsection{Contextual Bandits for Expert Selection}
\label{subsec:related_bandit}

Contextual bandits formalize sequential selection problems where an agent observes a context, chooses an action, and receives a reward~\cite{li2010contextual}. In the online setting, algorithms such as Thompson Sampling~\cite{thompson1933likelihood} and upper confidence bound~\cite{auer2002finite} balance exploration and exploitation with known regret guarantees~\cite{russo2018tutorial}. When online exploration is impractical, the offline (batch) bandit setting applies, where the policy is learned entirely from logged data without further interaction. A central principle in offline bandits is \textit{pessimism in the face of uncertainty}: the agent should avoid selecting actions whose predicted rewards are unreliable, preferring conservatively estimated outcomes~\cite{nguyen2021offline}. The LCB instantiates this principle by penalizing actions with high predictive variance. Scaling to high-dimensional contexts, the Neural Linear approach, which performs Bayesian inference with learned features, achieves the best trade-off between uncertainty quality and computational cost across bandit problems~\cite{riquelme2018deep}, directly motivating our VBLL-based design. We apply the offline bandit formulation to diffusion policy expert routing, using VBLL for reward prediction and LCB for pessimistic selection.

\section{PRELIMINARIES}
\label{sec:methodology}

Our methodology centers on learning a diffusion policy for multi-modal active target tracking with uncertainty-aware expert selection. We first describe the expert planners that generate demonstration data, then present the policy network architecture, consisting of an observation encoder and an expert-conditioned diffusion policy for action generation. The Bayesian expert selection mechanism is detailed in Section~\ref{sec:vbll}.

\subsection{Problem Formulation}
\label{subsec:problem}

Consider a mobile robot with state $\bx_t \in \Real^{n_x}$ and control input $\bu_t \in \Real^{n_u}$, evolving according to $\bx_{t+1} = f(\bx_t, \bu_t)$. The robot operates in an environment containing $N_y$ targets with individual states $\by_t^{(j)} \in \Real^{n_y}$ for $j \in \{1, \ldots, N_y\}$. Both the exact number of targets $N_y$ and their dynamics are unknown. The robot is equipped with a sensor providing a limited FoV $\calF(\bx_t) \subset \Real^3$, yielding measurements
\begin{equation}
\bz_t^{(j)} = \mathbf{H} \by_t^{(j)} + \boldsymbol{\eta}_t, \quad \text{if } \bp(\by_t^{(j)}) \in \calF(\bx_t),
\label{eq:sensor_model}
\end{equation}
where $\bp(\cdot)$ denotes the position component, and $\boldsymbol{\eta}_t \sim \calN(\mathbf{0}, \mathbf{R})$ is measurement noise with covariance $\mathbf{R} \in \Real^{n_z \times n_z}$.

The agent maintains Gaussian beliefs $\by_t^{(j)}|\bz_{0:t} \sim \calN(\bbmu_t^{(j)}, \bbSig_t^{(j)})$ via Kalman filtering. The filter update executes only when a target is detected within the FoV, while prediction continues regardless. 

\subsubsection{Tracking performance metric}

We quantify tracking performance using the negative log-likelihood (NLL)~\cite{pinto2021uncertainty} of the true target state under the belief distribution, i.e., the standard Gaussian NLL $\mathrm{NLL}_t^{(j)} = \frac{1}{2}[\ln|\bbSig_t^{(j)}| + (\by_t^{(j)} - \bbmu_t^{(j)})^\top (\bbSig_t^{(j)})^{-1} (\by_t^{(j)} - \bbmu_t^{(j)}) + n_y \ln(2\pi)]$. The aggregate NLL over all existing targets is: 
\begin{equation}
\mathrm{NLL}_t = \frac{1}{|\calJ_t|} \sum_{j \in \calJ_t} \mathrm{NLL}_t^{(j)},
\label{eq:nll_aggregate}
\end{equation}
where $\calJ_t$ denotes the set of existing targets at time $t$. Lower NLL indicates better tracking performance, reflecting both accurate state estimation and confident predictions.

\subsubsection{Receding horizon control with adaptive expert selection}

We formulate the control problem in a receding-horizon framework. At each planning step $\tau = 0, T, 2T, \ldots$, the agent: 
\begin{enumerate}
\item[s1)] Observes the recent history $\mathbf{o}_{\tau-T_{\mathrm{obs}}+1:\tau}$ consisting of egocentric maps and target beliefs over the past $T_{\mathrm{obs}}$ steps;
\item[s2)] Selects an expert policy $\pi_{k_\tau}$ from $K$ available experts based on the current context;
\item[s3)] Generates an action sequence $\mathbf{a}_{\tau:\tau+T_{\mathrm{pred}}-1}$ of length $T_{\mathrm{pred}}$ conditioned on the selected expert;
\item[s4)] Executes only the first $T_{\mathrm{act}}$ actions before re-planning, constituting an open-loop feedback control structure~\cite{tse1973information}.
\end{enumerate}

This receding horizon structure allows the agent to adaptively switch between behavioral modes (exploration, tracking, reacquisition) as the context evolves. The expert selection is performed every $T_{\mathrm{act}}$ steps, enabling dynamic adaptation to changing target configurations and uncertainty levels throughout the trajectory.

\subsection{Expert Planners for Multi-Modal Demonstrations}
\label{subsec:experts}

We design three expert planners that generate a diverse dataset of trajectories covering different behavioral modes. The \textit{exploration planner} ($\pi_1$) performs frontier-based exploration~\cite{yamauchi1997frontier}, selecting frontier points based on distance, visitation frequency, and expected coverage gain. The \textit{uncertainty-based hybrid planner} ($\pi_2$) switches between exploration and tracking: when any detected target's covariance exceeds a threshold, it navigates toward the most uncertain target via Rapidly-exploring Random Tree star (RRT*)~\cite{karaman2011sampling}; otherwise, it explores. The \textit{time-based hybrid planner} ($\pi_3$) explores until detecting a target, tracks it for a fixed time interval, then resumes exploration. The demonstration dataset $\calD_N := \{(\mathbf{o}_n, \mathbf{a}_n, k_n)\}_{n=1}^N$ is collected by running these experts in simulation, where $\mathbf{o}_n$ denotes the observation, $\mathbf{a}_n$ is the action sequence, and $k_n \in \{1, 2, 3\}$ refers to the expert index.

\subsection{Observation Encoder}
\label{subsec:architecture}

The observation encoder maps heterogeneous inputs to a fixed-size feature vector. An egocentric occupancy-grid map is processed by a CNN followed by a Performer transformer~\cite{choromanski2021rethinking} to produce a map embedding $\mathbf{z}_{\mathrm{map}} \in \Real^{d_{\mathrm{map}}}$. Target beliefs $(\bbmu_t^{(j)}, \bbSig_t^{(j)})$, transformed into the robot's frame, are encoded via a self-attention module with masking for undetected targets, yielding $\mathbf{z}_{\mathrm{target}} \in \Real^{d_{\mathrm{target}}}$. The global feature representation is:
\begin{equation}
\bbphi_{\bf W}(\mathbf{o}) = [\mathbf{z}_{\mathrm{map}}^\top, \mathbf{z}_{\mathrm{target}}^\top]^\top \in \Real^d, \quad d = d_{\mathrm{map}} + d_{\mathrm{target}}.
\label{eq:feature_concat}
\end{equation}
In our implementation, $d_{\mathrm{map}} = d_{\mathrm{target}} = 256$, yielding $d = 512$. This shared encoder $\bbphi$ provides features for both the VBLL expert selector and the diffusion policy.

\subsection{Diffusion Policy with Expert Conditioning}
\label{subsec:diffusion}

The action generation is performed by a Denoising Diffusion Probabilistic Model (DDPM)-based diffusion policy~\cite{chi2025diffusion,ho2020denoising} conditioned on both the observation features and the selected expert. Specifically, the conditioning vector incorporates expert information via a learned embedding 
\begin{equation}
\mathbf{c}_n = [\bbphi_{\bf W}^\top (\mathbf{o}_n), \bbphi_{\bf E}^\top ({k}_n)]^\top \in \Real^{d + d_e}, \label{eq:c_n}
\end{equation}
where $\bbphi_{\bf E}^\top (\mathbf{k}_n)\in \Real^{d_e}$ is a feature extractor with learnable parameters ${\bf E}$ for expert $\mathbf{k}_n$. This allows the same diffusion backbone to generate qualitatively different action sequences depending on the selected expert, while sharing the observation encoder $\bbphi$ across all modes.

\textbf{Forward process.} Given a ground-truth action sequence $\mathbf{a}^{(0)}_n := \mathbf{a}_n$ from the demonstration dataset, the forward process progressively corrupts it over $I_{\mathrm{diff}}$ steps according to a variance schedule $\{\alpha_i\}_{i=1}^{I_{\mathrm{diff}}}$. The noisy action at step $n$ can be sampled in closed form
\begin{equation}
\mathbf{a}^{(i)}_n = \sqrt{\bar{\alpha}_i}\, \mathbf{a}^{(0)}_n + \sqrt{1 - \bar{\alpha}_i}\, \boldsymbol{\epsilon}_n^{(i)}, \quad \boldsymbol{\epsilon}_n^{(i)} \sim \calN(\mathbf{0}, \mathbf{I}),
\label{eq:forward}
\end{equation}
where $\bar{\alpha}_i = \prod_{s=1}^{i} \alpha_s$.

\textbf{Training procedure.} A U-Net noise predictor $\epsilon_\theta$ takes as input the noisy action $\mathbf{a}^{(i)}_n$, the diffusion timestep $i$, and the conditioning vector $\mathbf{c}_n$, and outputs a predicted noise $\hat{\boldsymbol{\epsilon}}_n^{(i)} = \epsilon_\theta(\mathbf{a}^{(i)}_n, i, \mathbf{c}_n)$. Given ${\cal D}_N$, the goal is to jointly optimize $\{\theta, {\bf W},{\bf E} \}$ by minimizing 
\begin{equation}
\mathcal{L}_{\mathrm{diff}}(\theta, {\bf W},{\bf E}) = \sum_{n=1}^N\sum_{i=1}^{I_{\rm diff}} \left[ \| \boldsymbol{\epsilon}_n^{(i)} - \epsilon_\theta(\mathbf{a}_n^{(i)}, i, \mathbf{c}_n) \|^2 \right],
\label{eq:diffusion_loss}
\end{equation}
where $\mathbf{a}^{(i)}_n$ is obtained via~\eqref{eq:forward}, and $\mathbf{c}_n$ is constructed from~\eqref{eq:c_n}. The U-Net parameters $\theta$, expert embeddings $\{\mathbf{e}_k\}_{k=1}^K$, and encoder parameters $\bbphi$ are updated jointly. 

\textbf{Inference procedure.} At deployment, the action sequence is generated by iteratively denoising from pure Gaussian noise. Starting from $\mathbf{a}^{(I_{\mathrm{diff}})} \sim \calN(\mathbf{0}, \mathbf{I})$, the reverse process applies the trained noise predictor at each step:
\begin{equation}
\mathbf{a}^{(i-1)} = \frac{1}{\sqrt{\alpha_i}} \left( \mathbf{a}^{(i)} - \frac{1 - \alpha_i}{\sqrt{1 - \bar{\alpha}_i}}\, \epsilon_\theta(\mathbf{a}^{(i)}, i, \mathbf{c}_\tau) \right) + \sigma_i \mathbf{z},
\label{eq:reverse}
\end{equation}
where $\mathbf{z} \sim \calN(\mathbf{0}, \mathbf{I})$ for $i > 1$ and $\mathbf{z} = \mathbf{0}$ for $i = 1$, and $\sigma_i^2 = \frac{(1 - \bar{\alpha}_{i-1})}{(1 - \bar{\alpha}_i)}(1 - \alpha_i)$ is the posterior variance of the reverse step with $\bar{\alpha}_0 := 1$~\cite{ho2020denoising}. After $I_{\mathrm{diff}}$ denoising steps, the output $\mathbf{a}^{(0)}$ is the generated action sequence.

When the expert embedding $\mathbf{e}_k$ is removed, the policy reduces to the unconditioned MATT-Diff baseline~\cite{liu2025matt}. In our experiments, all learning-based selection methods (fixed-expert, random, MoE, and VBLL) share the same expert-conditioned checkpoint and differ only in how $k$ is chosen at inference, isolating the effect of expert selection from architectural differences. The key remaining question, how to select $k$ at each re-planning step, is addressed next.

\section{OFFLINE CONTEXTUAL BANDIT FOR EXPERT SELECTION}
\label{sec:vbll}

We formulate expert selection as an \textit{offline contextual bandit} problem~\cite{li2010contextual,nguyen2021offline} embedded within the receding-horizon control framework: at each re-planning step, the agent observes a context (current observation) and selects an arm (expert), aiming to maximize tracking reward. We adopt VBLL~\cite{brunzema2024bayesian} for uncertainty-aware reward prediction and a LCB criterion for pessimistic selection.

\subsection{Sequential Expert Selection as Contextual Bandit}

At each re-planning step $\tau$, the agent faces a contextual bandit problem with context $\bbphi_{\widehat{\bf W}}(\mathbf{o}_\tau) \in \Real^d$ (the encoded observation history), $K$ arms corresponding to expert policies $\{\pi_1, \ldots, \pi_K\}$, and a reward model that predicts the expected tracking reward (negative NLL via Eq.~\eqref{eq:nll_aggregate}) of each expert given the context. Since the reward model is trained offline and held fixed at deployment, the agent cannot collect new feedback to refine its predictions. The pessimism principle is therefore essential: by preferring experts whose predicted rewards are both high and confident, LCB avoids overcommitting to experts in belief-state regions with sparse training coverage and unreliable predictions.

\subsection{Multi-Head VBLL Formulation}

To proceed, we define ${\cal N}_k:=\{n:k_n=k,k\in {\cal N}\}$ (${\cal N}:=\{1,\ldots,N\}$), which collects the training sample indices in ${\cal D}_N$ corresponding to expert $k\in \{1, \ldots, K\}$. We maintain a separate VBLL head that predicts the expected tracking reward when selecting that expert. We define the reward as the negative average NLL over the next $T$ steps
\begin{equation}
r^k_{n} = -\frac{1}{T} \sum_{t=nT}^{ (n+1)T} \mathrm{NLL}_t,
\label{eq:reward_target}
\end{equation}
where higher values indicate better tracking performance. Each head is formulated as a Bayesian last-layer regression model:
\begin{equation}
r_n^k = \bbphi_{\widehat{\bf W}}^\top (\mathbf{o}_n) \bbbeta^k + \eta_n^k, \quad \eta_n^k \sim \calN(0, \sigma_{\eta,k}^2),  \quad n\in{\cal N}_k
\label{eq:vbll_model}
\end{equation}
where ${\cal N}_k:=\{n:k_n=k,k\in {\cal N}\}$ (${\cal N}:=\{1,\ldots,N\}$), $r_n^k$ represents the predicted reward for expert $k$, $\bbphi_{\widehat{\bf W}}(\mathbf{o}_n) \in \Real^d$ is the feature vector from the frozen encoder (Eq.~\eqref{eq:feature_concat}), and $\bbbeta^k \in \Real^d$ is the random regression head with Gaussian prior $p(\bbbeta^k) = \calN(\bbbeta^k; \mathbf{0}, \sigma_\beta^2 \mathbf{I}_d)$.

\subsection{Variational Training}

Let the variational posterior of $\bbbeta^k$ be $q(\bbbeta^k) = \calN(\bbbeta^k; \bbmu_N^k, \bbSig_N^k)$. The posterior parameters $\{\bbmu_N^k, \bbSig_N^k\}$ are learned jointly by maximizing the evidence lower bound (ELBO):
\begin{align}
\mathcal{L}^{\mathrm{ELBO}}(\boldsymbol{\Theta}_k) &= \mathbb{E}_{q(\bbbeta^k)} \left[\sum_{n\in {\cal N}_k} \log p(r_n^k | \bbbeta^k, \bbphi_{\widehat{\bf W}} (\mathbf{o}_n))\right] \nonumber \\
&\quad - \mathrm{KL}(q(\bbbeta^k) \| p(\bbbeta^k)),
\label{eq:elbo}
\end{align}
where $\boldsymbol{\Theta}_k := \{\bbmu_N^k, \bbSig_N^k, \sigma_{\eta,k}^2\}$, 
and the ELBO admits a closed-form expression
\begin{align}
&\mathcal{L}^{\mathrm{ELBO}}(\boldsymbol{\Theta}_k) = -\frac{1}{2}\log(2\pi\sigma_{\eta,k}^2)  \nonumber\\
&- \frac{1}{2\sigma_\beta^2} \left( \mathrm{Tr}(\bbSig_N^k) + (\bbmu_N^k)^\top \bbmu_N^k \right)\nonumber \\
&- \frac{1}{2\sigma_{\eta,k}^2} \sum_{n\in{\cal N}_k} \left[ (r_n^k - \bbphi_{\widehat{\bf W}}^\top (\mathbf{o}_n)\bbmu_N^k)^2 \right]\nonumber \\
& - \frac{1}{2\sigma_{\eta,k}^2} \sum_{n\in{\cal N}_k} \bbphi_{\widehat{\bf W}}^\top (\mathbf{o}_n)\bbSig_N^k\bbphi_{\widehat{\bf W}}(\mathbf{o}_n) + \frac{1}{2} \left[ d + \log \frac{|\bbSig_N^k|}{\sigma_\beta^{2d}} \right].
\label{eq:elbo_explicit}
\end{align}
\textbf{Training procedure.} All diffusion policy parameters $(\theta, \bbphi, \{\mathbf{e}_k\})$ are frozen after the first stage (Section~\ref{subsec:diffusion}); only the VBLL parameters $\{\boldsymbol{\Theta}_k\}_{k=1}^K$ are updated. Each head $k$ is trained exclusively on samples from the corresponding expert $\pi_k$, ensuring specialization. The total loss aggregates the negative ELBO across all heads, weighted by sample proportion in each mini-batch of size $B$ 
\begin{equation}
\mathcal{L}_{\mathrm{VBLL}} (\{\boldsymbol{\Theta}_k\}_k)  = \sum_{k=1}^{K} \frac{|{\cal B}_k|}{B} \cdot \bigl(-\mathcal{L}^{\mathrm{ELBO}}(\boldsymbol{\Theta}_k)\bigr),
\label{eq:vbll_loss}
\end{equation}
where ${\cal B}_k:=\{k_n=k: k_n\in{\cal B}\} $ collects samples contributing to head $k$'s ELBO (Eq.~\eqref{eq:elbo_explicit}).

\subsection{Predictive Distribution}

Given the learned posterior $q(\bbbeta^k) = \calN(\bbbeta^k; \hat{\bbmu}^k_N, \hat{\bbSig}_N^k)$, the predictive distribution for expert $k$ at a new observation $\mathbf{o}_*$ is:
\begin{equation}
p(r^k | \mathbf{o}_*, \calD_N) = \calN(r_*^k; \hat{r}_*^k, (\sigma_*^k)^2),
\label{eq:predictive}
\end{equation}
where the predictive mean and variance are:
\begin{align}
\hat{r}_*^k &= \bbphi_{\widehat{\bf W}}^\top(\mathbf{o}_*) \hat{\bbmu}_N^k, \label{eq:pred_mean} \\
(\sigma_*^k)^2 &= \bbphi_{\widehat{\bf W}}^\top(\mathbf{o}_*) \hat{\bbSig}_N^k \bbphi_{\widehat{\bf W}}(\mathbf{o}_*) + \hat{\sigma}_{\eta,k}^2.
\label{eq:pred_var}
\end{align}

\subsection{Expert Selection via Lower Confidence Bound}

At each re-planning step $\tau$, we perform expert selection following the pessimism principle for offline decision-making~\cite{nguyen2021offline}. Given the predictive distribution in Eq.~\eqref{eq:predictive}, we compute a LCB on the predicted reward for each expert 
\begin{equation}
\mathrm{LCB}_*^k = \hat{r}^k_* - \lambda \cdot \sigma^k_*, \quad k = 1, \ldots, K,
\label{eq:lcb_score}
\end{equation}
where $\lambda > 0$ is a pessimism coefficient that controls the penalty for predictive uncertainty. Subtracting $\lambda \cdot \sigma^k$ reduces the estimated reward for uncertain experts, implementing pessimism in the standard reward-maximization setting. The selected expert is
\begin{equation}
\hat{k}_* = \arg\max_{k \in \{1, \ldots, K\}} \mathrm{LCB}_*^k.
\label{eq:lcb_selection}
\end{equation}

This formulation embodies the pessimism principle standard in offline bandits and offline reinforcement learning~\cite{nguyen2021offline,swaminathan2015batch}: the reward model is learned from logged demonstration data and held fixed at deployment, so the agent cannot collect new feedback to correct overoptimistic predictions. LCB guards against this by preferring experts whose predicted rewards are both high \emph{and} confident. The coefficient $\lambda$ controls conservatism: a larger $\lambda$ favors well-understood experts. The Bayesian formulation further handles distribution mismatch: when context $\bbphi(\mathbf{o_*})$ lies outside the training support of head $k$, the epistemic uncertainty term in Eq.~\eqref{eq:pred_var} grows, lowering the LCB score and naturally routing selection toward experts with better coverage of the current belief state.

\subsection{Inference Procedure}
\label{subsec:inference}

Algorithm~\ref{alg:inference} summarizes the complete receding horizon control procedure with VBLL-guided expert selection. At each re-planning step (every $T_{\mathrm{act}}$ time steps), the agent extracts features from the recent observation history, computes the LCB score for each expert to perform pessimistic selection, generates an action sequence conditioned on the selected expert, and executes the first $T_{\mathrm{act}}$ actions before re-planning. This allows the agent to dynamically switch between exploration, tracking, and reacquisition modes as the tracking scenario evolves.

\begin{algorithm}[t]
\caption{Receding-horizon control with VBLL-LCB expert selection}
\label{alg:inference}
\begin{algorithmic}[1]
\REQUIRE Trained encoder $\bbphi_{\widehat{\bf W}}$, trained diffusion policy $\epsilon_\theta$ with expert embeddings $\bbphi_{\widehat{\bf E}}$, trained VBLL parameters $\{\hat{\boldsymbol{\Theta}}_k\}_{k=1}^K$, pessimism coefficient $\lambda$, variance schedule $\{\alpha_i\}_{i=1}^{I_{\mathrm{diff}}}$, horizons $T_{\mathrm{obs}}, T_{\mathrm{pred}}, T_{\mathrm{act}}$
\STATE Initialize observation buffer; set $\tau \leftarrow 0$
\WHILE{episode not terminated}
    \STATE Collect observation history $\mathbf{o}_{\tau-T_{\mathrm{obs}}+1:\tau}$
    \STATE Encode observation $\bbphi_{\widehat{\bf W}}(\mathbf{o}_\tau) \in \Real^d$ via~\eqref{eq:feature_concat}
    \FOR{$k = 1$ to $K$}
        \STATE Compute predictive mean $\hat{r}^k_\tau$ and variance $(\sigma^k_\tau)^2$ via~\eqref{eq:pred_mean}--\eqref{eq:pred_var}
        \STATE $\mathrm{LCB}^k_\tau \leftarrow \hat{r}^k_\tau - \lambda \cdot \sigma^k_\tau$
    \ENDFOR
    \STATE Select expert $\hat{k}_\tau \leftarrow \arg\max_{k \in \{1, \ldots, K\}} \mathrm{LCB}^k_\tau$
    \STATE Form conditioning $\mathbf{c}_\tau \leftarrow [\bbphi_{\widehat{\bf W}}^\top(\mathbf{o}_\tau),\; \bbphi_{\widehat{\bf E}}^\top(\hat{k}_\tau)]^\top \in \Real^{d+d_e}$ via~\eqref{eq:c_n}
    \STATE Sample $\mathbf{a}^{(I_{\mathrm{diff}})} \sim \calN(\mathbf{0}, \mathbf{I})$
    \FOR{$i = I_{\mathrm{diff}}$ \textbf{down to} $1$}
        \STATE Denoise $\mathbf{a}^{(i)} \to \mathbf{a}^{(i-1)}$ with conditioning $\mathbf{c}_\tau$ via~\eqref{eq:reverse}
    \ENDFOR
    \STATE Set $\mathbf{a}_{\tau:\tau+T_{\mathrm{pred}}-1} \leftarrow \mathbf{a}^{(0)}$
    \FOR{$t = \tau$ to $\tau + T_{\mathrm{act}} - 1$}
        \STATE Execute $\mathbf{a}_t$; update beliefs $\{(\bbmu_t^{(j)}, \bbSig_t^{(j)})\}$ via Kalman filter
    \ENDFOR
    \STATE $\tau \leftarrow \tau + T_{\mathrm{act}}$
\ENDWHILE
\end{algorithmic}
\end{algorithm}

\section{EXPERIMENTS}
\label{sec:experiments}

We design experiments to evaluate three aspects of our framework: (1)~whether explicit, uncertainty-aware expert selection improves tracking performance over both implicit mode selection and deterministic gating; (2)~whether the regression-based bandit formulation outperforms classification-based routing; and (3)~whether leveraging predictive uncertainty in the selection rule yields better decisions than ignoring it.

\subsection{Experimental Setup}
\label{subsec:setup}

\subsubsection{Environment and dataset}

We evaluate in a 2-D indoor floor plan from HouseExpo~\cite{li2020houseexpo}. Following the setup of \cite{liu2025matt}, we report results over 5 episodes with fixed random seeds for reproducibility. Each episode runs for 1{,}000 steps with $N_y \in \{3, \ldots, 6\}$ moving targets whose initial positions are randomized. The targets follow a Brownian velocity model~\cite{li2003survey} with process noise $\mathbf{Q} = \mathrm{diag}(90, 40)$. The robot observes targets through a limited FoV sensor with measurement model $\mathbf{H} = \mathbf{I}$ and noise $\mathbf{R} = \mathrm{diag}(0.05^2, 0.05^2)$, and maintains Gaussian beliefs via Kalman filtering with conservative noise estimates.

The demonstration dataset is collected from the three expert planners described in Section~\ref{subsec:experts}: frontier-based exploration ($\pi_1$), uncertainty-based hybrid ($\pi_2$), and time-based hybrid ($\pi_3$). Each sample is labeled with the generating expert identity $k \in \{1, 2, 3\}$ and the corresponding tracking NLL over the subsequent $T_{\mathrm{act}}$ steps (Eq.~\eqref{eq:reward_target}).

\subsubsection{Evaluation metrics}

Following~\cite{pinto2021uncertainty}, we report three complementary metrics computed over detected targets, each averaged across the episode:
\begin{itemize}
\item \textbf{RMSE}: Root mean squared error between estimated and true target positions, measuring estimation accuracy.
\item \textbf{NLL}: NLL of true target states under the belief distribution (Eq.~\eqref{eq:nll_aggregate}), capturing both accuracy and calibration.
\item \textbf{Entropy}: Differential entropy of the belief distributions, reflecting estimation confidence.
\end{itemize}
Lower values indicate better tracking performance. We report the mean $\pm$ standard deviation over 5 episodes with fixed seeds for reproducibility. To isolate the effect of expert selection from action generation, we evaluate under both \emph{rule-based} execution (where the selected expert's handcrafted planner generates actions) and \emph{learning-based} execution (where the expert-conditioned diffusion policy generates actions).

\subsection{Baselines and Comparisons}
\label{subsec:baselines}

Our baselines are organized to enable controlled comparisons that directly test the claims outlined above.

\noindent\textbf{Individual expert planners} serve as performance references for each behavioral mode: \textit{Track-rule} ($\pi_3$, prioritizes maintaining visibility of detected targets), \textit{Reacq-rule} ($\pi_2$, balances tracking with re-acquiring lost targets), and \textit{Explore-rule} ($\pi_1$, frontier-based exploration for coverage). \textit{Random-rule} selects uniformly among the three planners at each re-planning step.

\noindent\textbf{Implicit mode selection.} \textit{MATT-Diff}~\cite{liu2025matt} trains a single diffusion policy on the mixed demonstration dataset without expert labels. Mode selection is performed implicitly through the stochastic denoising process. This is the primary baseline our framework is designed to improve upon.

\noindent\textbf{Fixed expert conditioning.} \textit{DDPM-$\langle$Expert$\rangle$} trains an expert-conditioned diffusion policy but fixes the expert identity throughout the episode (Track, Reacq, or Explore). \textit{DDPM-Random} selects the expert uniformly at random at each re-planning step. These baselines isolate the value of context-aware selection: if adaptive selection provides no benefit, random selection should perform comparably.

\noindent\textbf{Deterministic gating methods.} To test whether uncertainty quantification in the selector matters, we compare against two point-estimate baselines that use the same encoder features but lack predictive uncertainty: \textit{MoE Selection} trains a softmax-based MLP classifier that maps the encoder features $\bbphi_{\widehat{\bf W}}(\mathbf{o})$ to expert probabilities and selects the most likely expert; \textit{MLP Regression} trains a deterministic MLP regressor that predicts expected NLL per expert and selects the expert with the lowest predicted NLL. Both represent the standard gating paradigm where routing outputs are point estimates.

\textbf{Alternative UQ methods.} To validate the choice of VBLL, we compare against \textit{MC Dropout}~\cite{gal2016dropout} ($T=20$ forward passes) and \textit{Neuralbandit}~\cite{nguyen2021offline}.

\textbf{Our methods.} \textit{VBLL} combines VBLL-based expert selection with LCB and diffusion policy for action generation. \textit{VBLL-rule} replaces the diffusion policy with the selected rule-based planner, isolating the expert selection mechanism from the action generation quality.

For fair comparison, all uncertainty-aware methods (VBLL, MC Dropout, and Neuralbandit) use the same LCB selection criterion (Eq.~\eqref{eq:lcb_score}) with $\lambda = 1$, isolating the effect of the uncertainty quantification method from the selection strategy.

\subsection{Main Results}
\label{subsec:results}

Table~\ref{tab:main_results} presents the tracking performance across all methods.

\begin{table}[t]
\centering
\caption{Tracking performance over 5 episodes. \textbf{Best} and 
\underline{second-best} results marked within each category.}
\label{tab:main_results}
\scriptsize
\setlength{\tabcolsep}{3pt}
\begin{tabular}{lccc}
\toprule
Method & RMSE $\downarrow$ & NLL $\downarrow$ & Entropy $\downarrow$ \\
\midrule
\multicolumn{4}{l}{\textit{Rule-based execution}} \\
Track-rule  & $499.96 \pm 154.04$ & $14.48 \pm 1.25$ & $13.41$ \\
Reacq-rule & $486.02 \pm 141.84$ & $14.43 \pm 1.11$ & $13.38$ \\
Explore-rule & $\underline{481.80 \pm 144.9}$ & $\underline{14.24 \pm 1.20}$ & $\underline{13.17}$ \\
Random-rule & $486.90 \pm 138.8$ & $14.31 \pm 1.05$ & $13.20$ \\
VBLL-rule (ours) & $\mathbf{472.90 \pm 129.14}$ & $\mathbf{13.83 \pm 1.13}$ & $\mathbf{12.78}$ \\
\midrule
\multicolumn{4}{l}{\textit{Learning-based execution: fixed \& deterministic}} \\
MATT-Diff & $521.57 \pm 140.34$ & $14.98\pm 1.11$ & $13.59$ \\
DDPM-Track & $554.46 \pm 36.22$ & $14.95 \pm 0.41$ & $13.48$ \\
DDPM-Reacq & $512.15 \pm 117.68$ & $14.82 \pm 0.93$ & $13.55$ \\
DDPM-Explore & $514.18 \pm 120.76$ & $14.64 \pm 1.04$ & $13.37$ \\
DDPM-Random & $524.30 \pm 113.2$ & $14.90 \pm 0.91$ & $13.53$ \\
DDPM-MoE Selection & $515.18 \pm 120.55$ & $14.86 \pm 1.03$ & $13.56$ \\
DDPM-MLP Regression & $511.33 \pm 126.64$ & $14.78\pm 1.01$ & $13.50$ \\
\midrule
\multicolumn{4}{l}{\textit{Learning-based execution: uncertainty-aware}} \\
DDPM-MC Dropout & $505.53 \pm 126.75$ & $14.98 \pm 1.03$ & $13.59$ \\
DDPM-Neuralbandit & $\underline{497.92 \pm 108.90}$ & $\underline{14.39 \pm 0.85}$ & $\underline{13.36}$ \\
VBLL (ours) & $\mathbf{480.80 \pm 107.27}$ & $\mathbf{14.06 \pm 0.89}$ & $\mathbf{13.01}$ \\
\bottomrule
\end{tabular}
\end{table}

\subsubsection{Explicit selection outperforms implicit mode selection}

The unconditioned MATT-Diff baseline relies on the stochastic denoising process for implicit mode selection. VBLL improves over this baseline across all three metrics, confirming that principled expert selection guided by the current belief state yields better mode choices than leaving strategy selection to randomness in denoising. Notably, DDPM-Random performs comparably to or slightly worse than the unconditioned baseline, indicating that the expert-conditioned diffusion policy is not inherently a stronger model. Rather, the improvement of VBLL originates entirely from the quality of expert selection, not from architectural differences in the action generator.

\subsubsection{Uncertainty-aware selection outperforms deterministic gating}

Both deterministic gating baselines, MoE Selection and MLP Regression, yield only marginal improvements over the unconditioned baseline, demonstrating that learned selection without uncertainty awareness provides limited benefit. Among the two, MLP Regression outperforms MoE Selection across all metrics, confirming that predicting per-expert rewards provides a richer training signal than classifying expert identity. VBLL further improves upon MLP Regression by leveraging predictive uncertainty through the LCB criterion, supporting our claim that both the regression-based bandit formulation and uncertainty-aware selection are essential for effective expert routing.

\subsubsection{Adaptive selection surpasses any fixed expert}

VBLL outperforms all fixed-expert variants and random selection in both rule-based and learning-based execution. The improvement over random selection confirms that gains stem from context-dependent selection rather than mere expert diversity. Additionally, VBLL narrows the gap between learned and rule-based execution to near parity, suggesting that, in prior diffusion-based tracking, the lack of informed mode selection was a larger source of error than the quality of action generation.

\begin{figure}[ht]
\centering
\includegraphics[width=1.0\columnwidth]{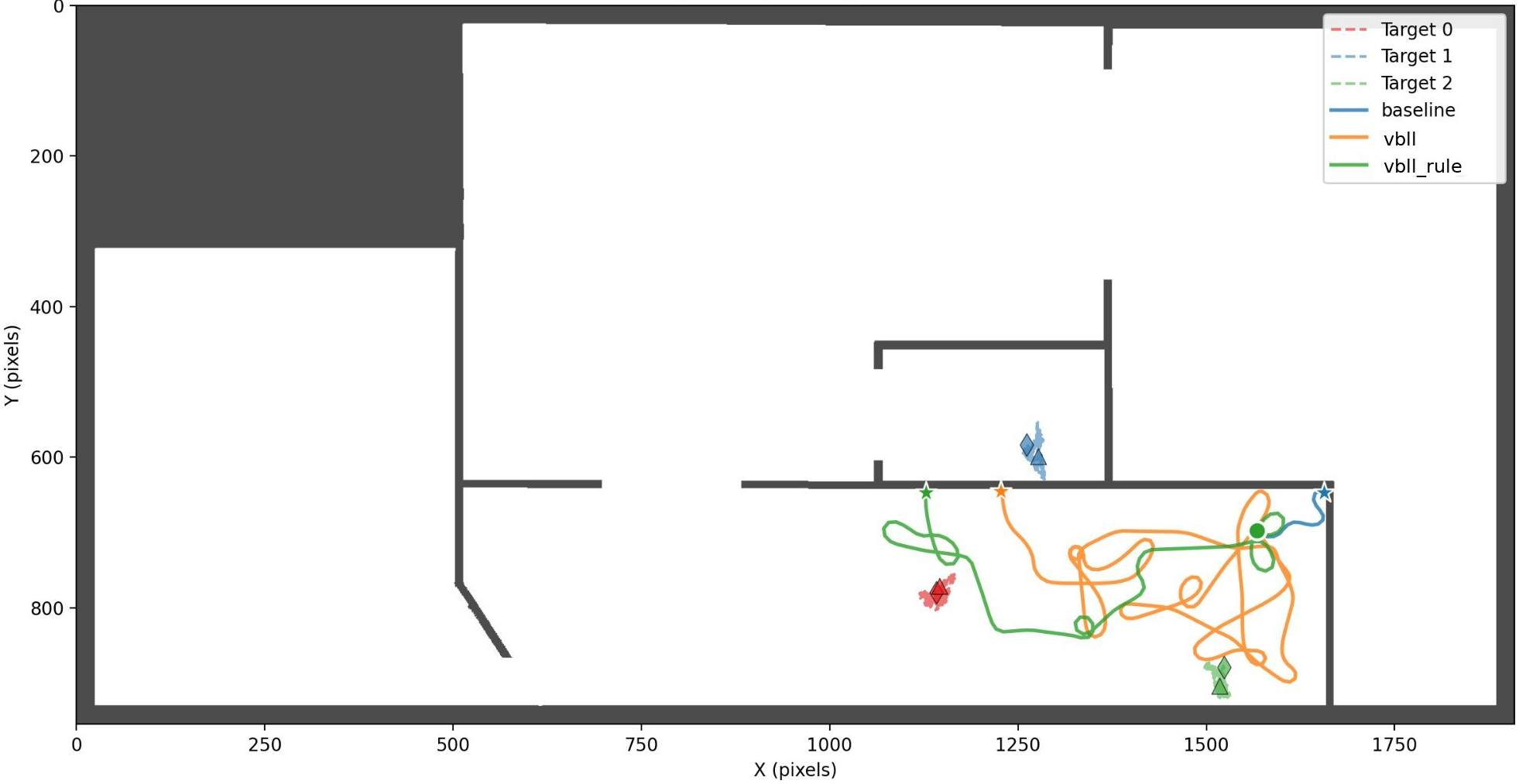}
\caption{Trajectory comparison on a representative episode with three targets (marked by colored diamonds).}
\label{fig:trajectory}
\end{figure}

Figure~\ref{fig:trajectory} illustrates the behavioral difference in a representative episode. The baseline policy (blue) explores a limited region, as the implicit mode selection during denoising does not consistently produce goal-directed behavior. The rule-based variant (green) efficiently navigates to target locations with compact trajectories. The learned policy (orange) covers a wider area but with notably less efficient paths, indicating that while VBLL expert selection meaningfully improves over the baseline, action generation quality remains a contributing factor to the performance gap between rule-based and learned execution.

\subsection{Ablation Studies}
\label{subsec:ablation}

\subsubsection{Selection strategy}

\begin{table}[t]
\centering
\caption{Ablation on expert selection strategy.}
\label{tab:ablation_strategy}
\scriptsize
\setlength{\tabcolsep}{3pt}
\begin{tabular}{lccc}
\toprule
Selection Strategy & RMSE $\downarrow$ & NLL $\downarrow$ & Entropy $\downarrow$ \\
\midrule
VBLL-Greedy (mean) & $500.29 \pm 102.79$ & $14.98 \pm 0.35$ & $13.48$ \\
VBLL-UCB & $491.01 \pm 141.84$ & $14.69 \pm 1.61$ & $13.44$ \\
VBLL-TS & $484.1 \pm 105.1$ & $14.35 \pm 0.79$ & $13.33$ \\
VBLL-LCB & $480.80 \pm 107.27$ & $14.06 \pm 0.89$ & $13.01$ 
 \\
\bottomrule
\end{tabular}
\end{table}

Table~\ref{tab:ablation_strategy} compares four selection strategies applied to the same VBLL model. Greedy selection, using only the predictive mean, performs comparably to the deterministic MLP Regression baseline (Table~\ref{tab:main_results}), confirming that accurate reward prediction alone is insufficient: without leveraging uncertainty, VBLL offers no advantage over a standard regressor. Incorporating uncertainty consistently improves performance, with LCB achieving the best results across all metrics. This aligns with the offline setting: since the reward model is fixed at deployment, pessimism is the appropriate principle, penalizing uncertain predictions to avoid overcommitment to experts with poor training coverage.

\subsubsection{Choice of UQ method}

Table~\ref{tab:main_results} further compares VBLL against alternative UQ methods, all using the same LCB criterion for selection. MC Dropout performs worst, consistent with known calibration limitations under distribution shift~\cite{ovadia2019can}. Neuralbandit achieves reasonable NLL owing to the conservative confidence widths produced by its NTK-based uncertainty estimation, but exhibits notably higher RMSE, suggesting less accurate reward prediction. VBLL outperforms both across all metrics while requiring only a single forward pass per expert, compared to $T=20$ for MC Dropout, a practical advantage for real-time control.

\subsubsection{Sensitivity to pessimism coefficient}

\begin{table}[t]
\centering
\caption{Sensitivity to pessimism coefficient $\lambda$ in LCB selection.}
\label{tab:beta_sensitivity}
\scriptsize
\setlength{\tabcolsep}{3pt}
\begin{tabular}{lccc}
\toprule
$\lambda$ & RMSE $\downarrow$ & NLL $\downarrow$ & Entropy $\downarrow$ \\
\midrule
0 (Greedy) & $500.29 \pm 102.79$ & $14.98 \pm 0.35$ & $13.48$ \\
0.1  & $490.55\pm117.29$ & $14.36\pm1.30$ & $13.19$ \\
1    & $480.80 \pm 107.27$ & $14.06 \pm 0.89$ & $13.01$\\
3    & $489.72\pm106.29$ & $14.03\pm0.93$ & $12.94$ \\
\bottomrule
\end{tabular}
\end{table}

Table~\ref{tab:beta_sensitivity} examines the sensitivity of tracking performance to the pessimism coefficient $\lambda$ in the LCB criterion (Eq.~\eqref{eq:lcb_score}). At $\lambda=0$, the selection reduces to greedy mean-only prediction, recovering the performance of VBLL + Greedy in Table~\ref{tab:ablation_strategy}. As $\lambda$ increases, the LCB criterion increasingly penalizes experts with high predictive uncertainty. Overall, we adopt $\lambda=1$ as a robust default that performs well across all three metrics.

\begin{table}[t]
\centering
\caption{Per-step inference time (ms)}

\label{tab:inference_time}
\scriptsize
\setlength{\tabcolsep}{3pt}
\begin{tabular}{lc}
\toprule
Method & Time (ms) $\downarrow$ \\
\midrule
Explore-rule       & 184.12  \\
Matt-Diff      & 15.21  \\
VBLL (ours)        & 16.25  \\
\bottomrule
\end{tabular}
\end{table}

\subsubsection{Computational efficiency}

Table~\ref{tab:inference_time} reports per-step inference time. The rule-based Explore planner relies on online RRT* path planning, resulting in roughly $10\times$ higher latency than the learning-based methods. Compared to the MATT-Diff baseline, VBLL introduces negligible overhead for expert selection while meaningfully improving tracking performance, indicating a favorable cost-performance trade-off.

\section{CONCLUSION}
\label{sec:conclusion}

We presented a Bayesian framework for uncertainty-aware expert selection in diffusion-based active multi-target tracking, formulating expert routing as an offline contextual bandit problem. By combining multi-head VBLL for uncertainty-aware, single-pass reward prediction with a LCB criterion for pessimistic selection, our approach decouples strategy selection from action generation and avoids overcommitment to experts with unreliable predictions. Experiments demonstrate consistent improvements over both the unconditioned MATT-Diff baseline and deterministic gating methods, with ablations confirming that leveraging predictive uncertainty through pessimism, rather than ignoring it or using it for exploration, is beneficial for effective offline expert routing. Future work includes scaling to larger expert sets, extending to the online bandit setting where the VBLL heads are updated from deployment experience to adapt to non-stationary tracking dynamics, improving the fidelity of learned action generation through more expressive generative models to close the remaining gap between rule-based and learned execution, and validation on physical robot platforms.

\bibliographystyle{IEEEtaes}
\bibliography{references}

@STRING{ICASSP = {Proc. IEEE Int. Conf. Acoust., Speech, Sig. Process.}}

@STRING{IEEESPL = {IEEE Sig. Process. Lett.}}

@STRING{ICML = {Proc. Int. Conf. Mach. Learn.} }

@STRING{ICLR = {Proc. Int. Conf. Learn. Represent.} }

@STRING{NIPS = {Proc. Adv.  Neural Inf. Process. Syst.}}

@article{gal2016dropout,
  title={Dropout as a {B}ayesian approximation: {R}epresenting model uncertainty in deep learning},
  author={Gal, Yarin and Ghahramani, Zoubin},
  journal = ICML,
  pages={1050--1059},
  year={2016},
}

@article{brunzema2024bayesian,
  title={{B}ayesian {O}ptimization via {C}ontinual {V}ariational {L}ast {L}ayer {T}raining},
  author={Brunzema, Paul and Jordahn, Mikkel and Willes, John and Trimpe, Sebastian and Snoek, Jasper and Harrison, James},
  journal= ICLR,
  year={2025}
}

@article{lakshminarayanan2017simple,
  title={{S}imple and {S}calable {P}redictive {U}ncertainty {E}stimation using {D}eep {E}nsembles},
  author={Lakshminarayanan, Balaji and Pritzel, Alexander and Blundell, Charles},
  journal= NIPS,
  volume={30},
  year={2017}
}

@article{harrison2024variational,
  title={{V}ariational {B}ayesian {L}ast {L}ayers},
  author={Harrison, James and Willes, John and Snoek, Jasper},
  journal=ICLR,
  year={2024}
}

@article{thompson1933likelihood,
  title={{O}n the {L}ikelihood that {O}ne {U}nknown {P}robability {E}xceeds {A}nother in {V}iew of the {E}vidence of {T}wo {S}amples},
  author={Thompson, William R},
  journal={Biometrika},
  volume={25},
  number={3/4},
  pages={285--294},
  year={1933},
  publisher={JSTOR}
}

@article{russo2018tutorial,
  title={A {T}utorial on {T}hompson {S}ampling},
  author={Russo, Daniel J and Van Roy, Benjamin and Kazerouni, Abbas and Osband, Ian and Wen, Zheng and others},
  journal={Foundations and Trends{\textregistered} in Machine Learning},
  volume={11},
  number={1},
  pages={1--96},
  year={2018},
  publisher={Now Publishers, Inc.}
}

@inproceedings{blundell2015weight,
  title={Weight uncertainty in neural network},
  author={Blundell, Charles and Cornebise, Julien and Kavukcuoglu, Koray and Wierstra, Daan},
  booktitle={International conference on machine learning},
  pages={1613--1622},
  year={2015},
  organization={PMLR}
}

@article{pinto2021uncertainty,
  title={An Uncertainty-Aware Performance Measure for Multi-Object Tracking},
  author={Pinto, Juliano and Xia, Yuxuan and Svensson, Lennart and Wymeersch, Henk},
  journal=IEEESPL,
  volume={28},
  pages={1689--1693},
  year={2021}
}

@inproceedings{li2020houseexpo,
  title={{HouseExpo}: A Large-Scale 2D Indoor Layout Dataset for Learning-Based Algorithms on Mobile Robots},
  author={Li, Siyu and Li, Zejian and Xiao, Jiadi and He, Yixian and Zhong, Binhui and Li, Zhiyong},
  booktitle={IEEE/RSJ International Conference on Intelligent Robots and Systems (IROS)},
  pages={5839--5846},
  year={2020}
}

@inproceedings{yamauchi1997frontier,
  title={A Frontier-Based Approach for Autonomous Exploration},
  author={Yamauchi, Brian},
  booktitle={IEEE International Symposium on Computational Intelligence in Robotics and Automation (CIRA)},
  pages={146--151},
  year={1997}
}

@article{karaman2011sampling,
  title={Sampling-Based Algorithms for Optimal Motion Planning},
  author={Karaman, Sertac and Frazzoli, Emilio},
  journal={International Journal of Robotics Research},
  volume={30},
  number={7},
  pages={846--894},
  year={2011}
}

@inproceedings{choromanski2021rethinking,
  title={Rethinking Attention with {P}erformers},
  author={Choromanski, Krzysztof M and Likhosherstov, Valerii and Dohan, David and Song, Xingyou and Gane, Andreea and Sarlos, Tamas and Hawkins, Peter and Davis, Jared Q and Mohiuddin, Afroz and Kaiser, Lukasz and Belanger, David B and Colwell, Lucy J and Weller, Adrian},
  booktitle=ICLR,
  year={2021}
}

@inproceedings{ho2020denoising,
  title={Denoising Diffusion Probabilistic Models},
  author={Ho, Jonathan and Jain, Ajay and Abbeel, Pieter},
  booktitle=NIPS,
  volume={33},
  pages={6840--6851},
  year={2020}
}

@inproceedings{lavalle1997motion,
  title={Motion strategies for maintaining visibility of a moving target},
  author={LaValle, Steven M and Gonz{\'a}lez-Banos, Hector H and Becker, Craig and Latombe, J-C},
  booktitle={Proceedings of international conference on robotics and automation},
  volume={1},
  pages={731--736},
  year={1997},
  organization={IEEE}
}

@inproceedings{le2009trajectory,
  title={On trajectory optimization for active sensing in Gaussian process models},
  author={Le Ny, Jerome and Pappas, George J},
  booktitle={Proceedings of the 48h IEEE Conference on Decision and Control (CDC) held jointly with 2009 28th Chinese Control Conference},
  pages={6286--6292},
  year={2009},
  organization={IEEE}
}

@article{hero2011sensor,
  title={Sensor management: Past, present, and future},
  author={Hero, Alfred O and Cochran, Douglas},
  journal={IEEE Sensors Journal},
  volume={11},
  number={12},
  pages={3064--3075},
  year={2011},
  publisher={IEEE}
}

@article{zhou2018resilient,
  title={Resilient active target tracking with multiple robots},
  author={Zhou, Lifeng and Tzoumas, Vasileios and Pappas, George J and Tokekar, Pratap},
  journal={IEEE Robotics and Automation Letters},
  volume={4},
  number={1},
  pages={129--136},
  year={2018},
  publisher={IEEE}
}

@article{schlotfeldt2018anytime,
  title={Anytime planning for decentralized multirobot active information gathering},
  author={Schlotfeldt, Brent and Thakur, Dinesh and Atanasov, Nikolay and Kumar, Vijay and Pappas, George J},
  journal={IEEE Robotics and Automation Letters},
  volume={3},
  number={2},
  pages={1025--1032},
  year={2018},
  publisher={IEEE}
}

@article{sun2024moving,
  title={Moving target tracking by unmanned aerial vehicle: A survey and taxonomy},
  author={Sun, Nianyi and Zhao, Jin and Shi, Qing and Liu, Chang and Liu, Peng},
  journal={IEEE Transactions on Industrial Informatics},
  volume={20},
  number={5},
  pages={7056--7068},
  year={2024},
  publisher={IEEE}
}

@inproceedings{jeong2021deep,
  title={Deep reinforcement learning for active target tracking},
  author={Jeong, Heejin and Hassani, Hamed and Morari, Manfred and Lee, Daniel D and Pappas, George J},
  booktitle={2021 IEEE International Conference on Robotics and Automation (ICRA)},
  pages={1825--1831},
  year={2021},
  organization={IEEE}
}

@inproceedings{yang2023policy,
  title={Policy Learning for Active Target Tracking over Continuous $ SE (3) $ Trajectories},
  author={Yang, Pengzhi and Koga, Shumon and Asgharivaskasi, Arash and Atanasov, Nikolay},
  booktitle={Learning for Dynamics and Control Conference},
  pages={64--75},
  year={2023},
  organization={PMLR}
}

@article{liu2025matt,
  title={MATT-Diff: Multimodal Active Target Tracking by Diffusion Policy},
  author={Liu, Saida and Atanasov, Nikolay and Koga, Shumon},
  journal={arXiv preprint arXiv:2511.11931},
  year={2025}
}

@article{chi2025diffusion,
  title={Diffusion policy: Visuomotor policy learning via action diffusion},
  author={Chi, Cheng and Xu, Zhenjia and Feng, Siyuan and Cousineau, Eric and Du, Yilun and Burchfiel, Benjamin and Tedrake, Russ and Song, Shuran},
  journal={The International Journal of Robotics Research},
  volume={44},
  number={10-11},
  pages={1684--1704},
  year={2025},
  publisher={Sage Publications Sage UK: London, England}
}

@article{pearce2023imitating,
  title={Imitating human behaviour with diffusion models},
  author={Pearce, Tim and Rashid, Tabish and Kanervisto, Anssi and Bignell, Dave and Sun, Mingfei and Georgescu, Raluca and Macua, Sergio Valcarcel and Tan, Shan Zheng and Momennejad, Ida and Hofmann, Katja and others},
  journal={arXiv preprint arXiv:2301.10677},
  year={2023}
}

@article{ze20243d,
  title={3d diffusion policy: Generalizable visuomotor policy learning via simple 3d representations},
  author={Ze, Yanjie and Zhang, Gu and Zhang, Kangning and Hu, Chenyuan and Wang, Muhan and Xu, Huazhe},
  journal={arXiv preprint arXiv:2403.03954},
  year={2024}
}

@inproceedings{sridhar2024nomad,
  title={Nomad: Goal masked diffusion policies for navigation and exploration},
  author={Sridhar, Ajay and Shah, Dhruv and Glossop, Catherine and Levine, Sergey},
  booktitle={2024 IEEE International Conference on Robotics and Automation (ICRA)},
  pages={63--70},
  year={2024},
  organization={IEEE}
}

@inproceedings{cao2025dare,
  title={Dare: Diffusion policy for autonomous robot exploration},
  author={Cao, Yuhong and Lew, Jeric and Liang, Jingsong and Cheng, Jin and Sartoretti, Guillaume},
  booktitle={2025 IEEE International Conference on Robotics and Automation (ICRA)},
  pages={11987--11993},
  year={2025},
  organization={IEEE}
}

@article{liu2024rdt,
  title={Rdt-1b: a diffusion foundation model for bimanual manipulation},
  author={Liu, Songming and Wu, Lingxuan and Li, Bangguo and Tan, Hengkai and Chen, Huayu and Wang, Zhengyi and Xu, Ke and Su, Hang and Zhu, Jun},
  journal={arXiv preprint arXiv:2410.07864},
  year={2024}
}

@article{team2024octo,
  title={Octo: An open-source generalist robot policy},
  author={Team, Octo Model and Ghosh, Dibya and Walke, Homer and Pertsch, Karl and Black, Kevin and Mees, Oier and Dasari, Sudeep and Hejna, Joey and Kreiman, Tobias and Xu, Charles and others},
  journal={arXiv preprint arXiv:2405.12213},
  year={2024}
}

@article{reuss2024efficient,
  title={Efficient diffusion transformer policies with mixture of expert denoisers for multitask learning},
  author={Reuss, Moritz and Pari, Jyothish and Agrawal, Pulkit and Lioutikov, Rudolf},
  journal={arXiv preprint arXiv:2412.12953},
  year={2024}
}

@article{zhou2024variational,
  title={Variational distillation of diffusion policies into mixture of experts},
  author={Zhou, Hongyi and Blessing, Denis and Li, Ge and Celik, Onur and Jia, Xiaogang and Neumann, Gerhard and Lioutikov, Rudolf},
  journal={Advances in Neural Information Processing Systems},
  volume={37},
  pages={12739--12766},
  year={2024}
}

@article{wang2024sparse,
  title={Sparse diffusion policy: A sparse, reusable, and flexible policy for robot learning},
  author={Wang, Yixiao and Zhang, Yifei and Huo, Mingxiao and Tian, Ran and Zhang, Xiang and Xie, Yichen and Xu, Chenfeng and Ji, Pengliang and Zhan, Wei and Ding, Mingyu and others},
  journal={arXiv preprint arXiv:2407.01531},
  year={2024}
}

@article{shazeer2017outrageously,
  title={Outrageously large neural networks: The sparsely-gated mixture-of-experts layer},
  author={Shazeer, Noam and Mirhoseini, Azalia and Maziarz, Krzysztof and Davis, Andy and Le, Quoc and Hinton, Geoffrey and Dean, Jeff},
  journal={arXiv preprint arXiv:1701.06538},
  year={2017}
}

@article{fedus2022switch,
  title={Switch transformers: Scaling to trillion parameter models with simple and efficient sparsity},
  author={Fedus, William and Zoph, Barret and Shazeer, Noam},
  journal={Journal of Machine Learning Research},
  volume={23},
  number={120},
  pages={1--39},
  year={2022}
}

@article{hendawy2023multi,
  title={Multi-task reinforcement learning with mixture of orthogonal experts},
  author={Hendawy, Ahmed and Peters, Jan and D'Eramo, Carlo},
  journal={arXiv preprint arXiv:2311.11385},
  year={2023}
}

@article{ovadia2019can,
  title={Can you trust your model's uncertainty? evaluating predictive uncertainty under dataset shift},
  author={Ovadia, Yaniv and Fertig, Emily and Ren, Jie and Nado, Zachary and Sculley, David and Nowozin, Sebastian and Dillon, Joshua and Lakshminarayanan, Balaji and Snoek, Jasper},
  journal={Advances in neural information processing systems},
  volume={32},
  year={2019}
}

@inproceedings{watson2021latent,
  title={Latent derivative Bayesian last layer networks},
  author={Watson, Joe and Lin, Jihao Andreas and Klink, Pascal and Pajarinen, Joni and Peters, Jan},
  booktitle={International Conference on Artificial Intelligence and Statistics},
  pages={1198--1206},
  year={2021},
  organization={PMLR}
}

@inproceedings{li2010contextual,
  title={A contextual-bandit approach to personalized news article recommendation},
  author={Li, Lihong and Chu, Wei and Langford, John and Schapire, Robert E},
  booktitle={Proceedings of the 19th international conference on World wide web},
  pages={661--670},
  year={2010}
}

@article{riquelme2018deep,
  title={Deep bayesian bandits showdown: An empirical comparison of bayesian deep networks for thompson sampling},
  author={Riquelme, Carlos and Tucker, George and Snoek, Jasper},
  journal={arXiv preprint arXiv:1802.09127},
  year={2018}
}

@article{swaminathan2015batch,
  title={Batch learning from logged bandit feedback through counterfactual risk minimization},
  author={Swaminathan, Adith and Joachims, Thorsten},
  journal={The Journal of Machine Learning Research},
  volume={16},
  number={1},
  pages={1731--1755},
  year={2015},
  publisher={JMLR. org}
}

@article{nguyen2021offline,
  title={Offline neural contextual bandits: Pessimism, optimization and generalization},
  author={Nguyen-Tang, Thanh and Gupta, Sunil and Nguyen, A Tuan and Venkatesh, Svetha},
  journal={arXiv preprint arXiv:2111.13807},
  year={2021}
}

@article{sudha2026informative,
  title={An informative planning framework for target tracking and active mapping in dynamic environments with asvs},
  author={Sudha, Sanjeev Ramkumar and Popovi{\'c}, Marija and Coates, Erlend M},
  journal={IEEE Robotics and Automation Letters},
  volume={11},
  number={3},
  pages={2690--2697},
  year={2026},
  publisher={IEEE}
}

@article{lew2025aid,
  title={AID: Agent Intent from Diffusion for Multi-Agent Informative Path Planning},
  author={Lew, Jeric and Cao, Yuhong and Tan, Derek Ming Siang and Sartoretti, Guillaume},
  journal={arXiv preprint arXiv:2512.02535},
  year={2025}
}

@article{auer2002finite,
  title={Finite-time analysis of the multiarmed bandit problem},
  author={Auer, Peter and Cesa-Bianchi, Nicolo and Fischer, Paul},
  journal={Machine learning},
  volume={47},
  number={2},
  pages={235--256},
  year={2002},
  publisher={Springer}
}

@article{bostrom2021sensor,
  title={Sensor management for search and track using the Poisson multi-Bernoulli mixture filter},
  author={Bostr{\"o}m-Rost, Per and Axehill, Daniel and Hendeby, Gustaf},
  journal={IEEE Transactions on Aerospace and Electronic Systems},
  volume={57},
  number={5},
  pages={2771--2783},
  year={2021},
  publisher={IEEE}
}

@article{bostrom2021pmbm,
  title={PMBM filter with partially grid-based birth model with applications in sensor management},
  author={Bostr{\"o}m-Rost, Per and Axehill, Daniel and Hendeby, Gustaf},
  journal={IEEE Transactions on Aerospace and Electronic Systems},
  volume={58},
  number={1},
  pages={530--540},
  year={2021},
  publisher={IEEE}
}

@article{ragi2013uav,
  title={UAV path planning in a dynamic environment via partially observable Markov decision process},
  author={Ragi, Shankarachary and Chong, Edwin KP},
  journal={IEEE Transactions on Aerospace and Electronic Systems},
  volume={49},
  number={4},
  pages={2397--2412},
  year={2013},
  publisher={IEEE}
}

@article{zhang2025cooperative,
  title={Cooperative Dynamic Target Tracking: Distributed Time-Varying Optimization for Multi-UAV System},
  author={Zhang, Boyang and Hou, Yueqi and Yin, Hang and Lv, Maolong and Yang, Aiwu and Wu, Lirong},
  journal={IEEE Transactions on Aerospace and Electronic Systems},
  year={2025},
  publisher={IEEE}
}

@article{duan2020dynamic,
  title={Dynamic discrete pigeon-inspired optimization for multi-UAV cooperative search-attack mission planning},
  author={Duan, Haibin and Zhao, Jianxia and Deng, Yimin and Shi, Yuhui and Ding, Xilun},
  journal={IEEE Transactions on Aerospace and Electronic Systems},
  volume={57},
  number={1},
  pages={706--720},
  year={2020},
  publisher={IEEE}
}

@article{li2003survey,
  title={Survey of maneuvering target tracking. Part I. Dynamic models},
  author={Li, X Rong and Jilkov, Vesselin P},
  journal={IEEE Transactions on aerospace and electronic systems},
  volume={39},
  number={4},
  pages={1333--1364},
  year={2003},
  publisher={IEEE}
}

@article{xiang2025fine,
  title={Fine-tuning LLMs with variational Bayesian last layer for high-dimensional Bayesian optimization},
  author={Xiang, Haotian and Xu, Jinwen and Lu, Qin},
  journal={arXiv preprint arXiv:2510.01471},
  year={2025}
}

@inproceedings{xiang2026scalable,
  title     = {Scalable Bayesian Fine-Tuning of LLMs for Multi-Objective Bayesian Optimization},
  author    = {Xiang, Haotian and Zhang, Heng and Lu, Qin},
  booktitle = {Proceedings of the IEEE International Conference on Acoustics, Speech and Signal Processing (ICASSP)},
  year      = {2026},
  pages     = {},
  address   = {Barcelona, Spain},
  month     = {May},
  organization = {IEEE}
}

@inproceedings{tse1973information,
  title={Information patterns and classes of stochastic control laws},
  author={Tse, E and Bar-Shalom, Y},
  booktitle={1973 IEEE Conference on Decision and Control including the 12th Symposium on Adaptive Processes},
  pages={43--46},
  year={1973},
  organization={IEEE}
}

\begin{IEEEbiography}[{\includegraphics[width=1in,height=1.25in]{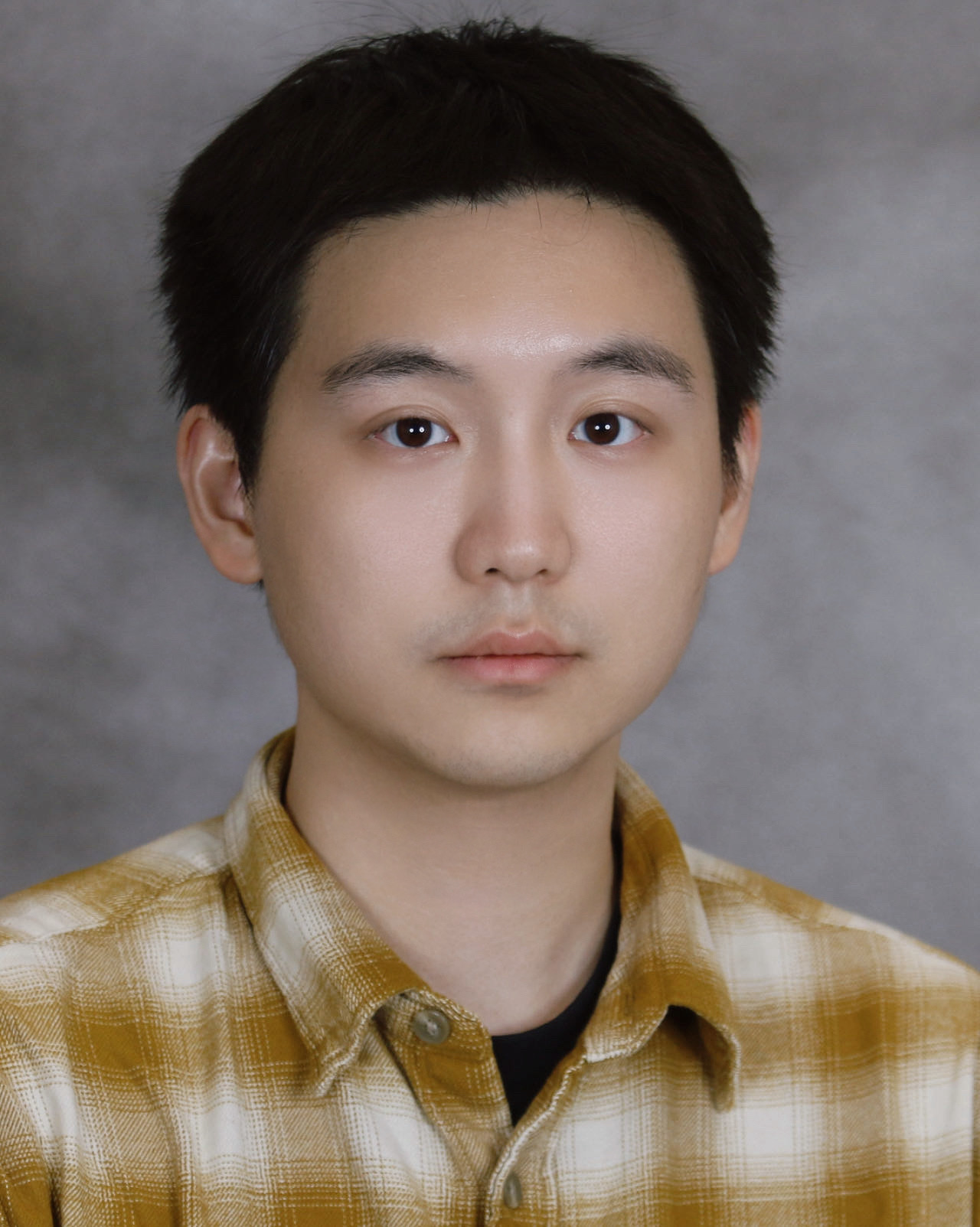}}]{Haotian Xiang}(Student Member, IEEE) received the B.S. degree in electrical engineering from the University of Electronic Science and Technology of China in 2022, and the M.S. degree in electrical engineering from Columbia University in 2023. He is currently pursuing the Ph.D. degree in electrical and computer engineering at the University of Georgia, under the supervision of Dr. Qin Lu. His research interests include Bayesian optimization, parameter-efficient fine-tuning, and uncertainty quantification for large language models.
\end{IEEEbiography}

\begin{IEEEbiography}[{\includegraphics[width=1in,height=1.25in]{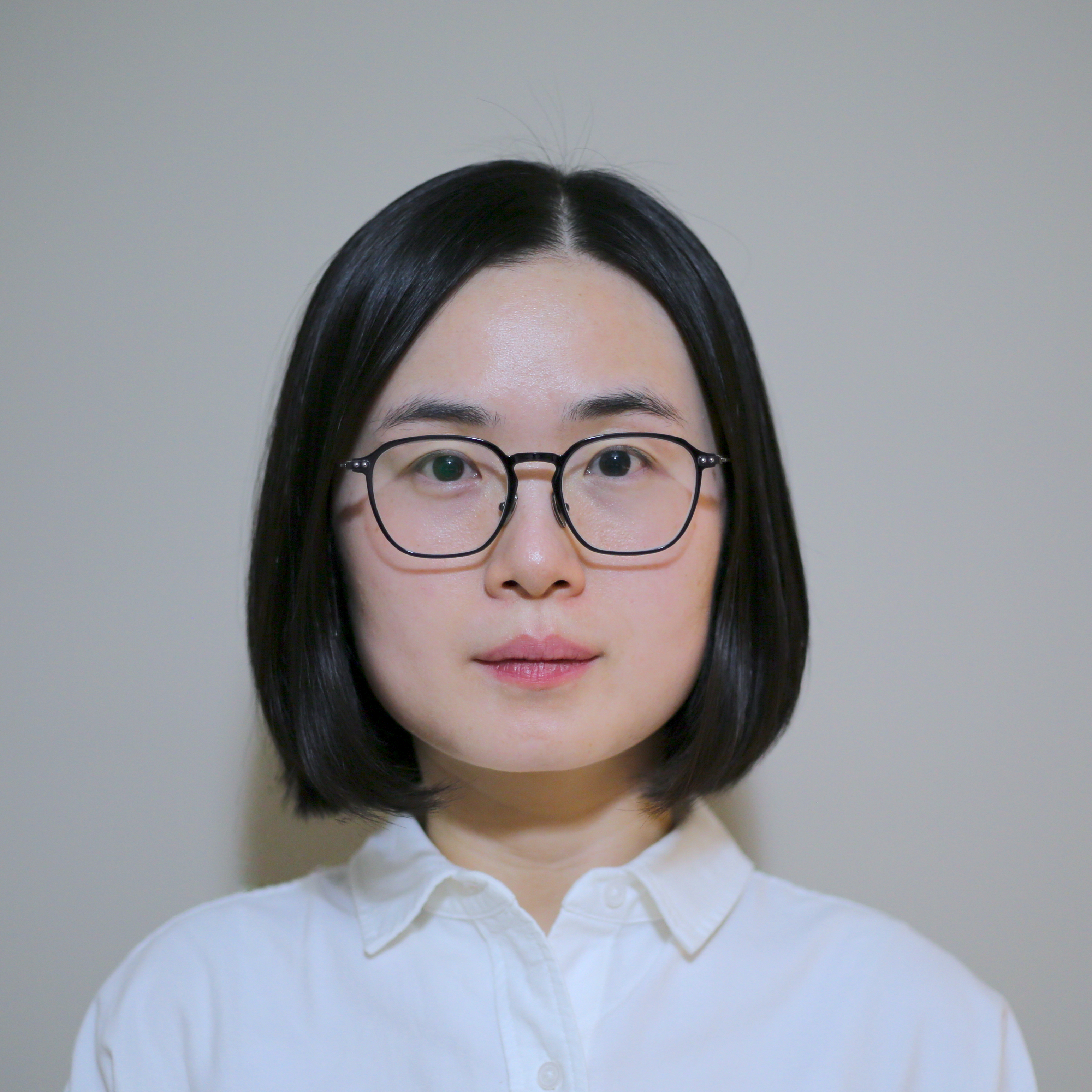}}]{Qin Lu}(Member, IEEE) received the B.S. degree from the University of Electronic Science and Technology of China in 2013 and the Ph.D. degree from the University of Connecticut (UConn) in 2018. Following the post-doctoral training at the University of Minnesota, she joined the School of Electrical and Computer Engineering at the University of Georgia as an Assistant Professor in 2023. Her research interests are in the areas of signal processing, machine learning, data science, and communications, with special focus on Gaussian processes, Bayesian optimization, spatio-temporal inference over graphs, and data association for multi-object tracking. She was awarded the Summer Fellowship and Doctoral Dissertation Fellowship from UConn. She was also a recipient of the Women of Innovation Award by Connecticut Technology Council in 2018, the NSF CAREER Award in 2024, Best Student Paper Award in IEEE Sensor Array and Multichannel Workshop in 2024, and the UConn Engineering GOLD Rising Star Alumni Award in 2025.
\end{IEEEbiography}

\begin{IEEEbiography}[{\includegraphics[width=1in,height=1.25in]{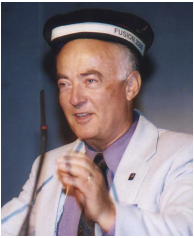}}]{Yaakov Bar-Shalom} (Life Fellow, IEEE) received the B.S. and M.S. degrees from the Technion—Israel Institute of Technology, Haifa, Israel, 1963 and 1967, respectively, and the Ph.D. degree from Princeton University, Princeton, NJ, USA, in 1970, all in electrical engineering.
He is currently a Board of Trustees Distinguished Professor with the Department of Electrical and Computer Engineering. and a M. E. Klewin Professor with the University of
Connecticut (UConn), Mansfield, CT, USA. His current research interests are in estimation theory, target tracking, and data fusion. He has authored or coauthored more than 650 papers and book chapters in these areas and in stochastic adaptive control and eight books, including Estimation with Applications to Tracking and Navigation (Wiley 2001) and Tracking and Data Fusion (2011). He is currently an Associate Editor for IEEE TRANSACTIONS ON AUTOMATIC CONTROL and Automatica, General Chairman of 1985 ACC, FUSION 2000, and was ISIF President (2000, 2002) and VP Publications (2004–2013). He graduated 42 Ph.D.s at UConn and served as Co-major advisor for 6 Ph.D. degrees awarded elsewhere. He is co-recipient of the M. Barry Carlton Award for the best paper in IEEE TRANSACTIONS ON AEROSPACE AND ELECTRONIC SYSTEMS (1995, 2000), the 2022 IEEE Aerospace and Electronic Systems Society Pioneer Award and recipient of the 2008 IEEE Dennis J. Picard Medal for Radar Technologies and Applications and the 2012 Connecticut Medal of Technology. He has been listed by academic.research.microsoft as \#1 in Aerospace Engineering based on the citations of his work and is the recipient of the 2015 ISIF Award for a Lifetime of Excellence in Information Fusion, renamed in 2016 as ``ISIF Yaakov Bar-Shalom Award for Lifetime of Excellence in Information Fusion." He is also recipient (with H.A.P. Blom) of the 2022 IEEE AESS Pioneer Award for the IMM Estimator. As of 2025, he graduated $44$ PhD students and served as co-major advisor for $6$ more at other institutions; $17$ of them are tenured or tenure-track professors.
\end{IEEEbiography}

\end{document}